\documentclass[acmsmall,nonacm,screen,11pt]{acmart}

\setcopyright{none}
\acmYear{2026}

\usepackage{brand}
\usepackage{header}
\usepackage{docspec}

\usepackage{amsmath}
\usepackage{dsfont}            
\usepackage{graphicx}
\usepackage{enumitem}
\usepackage{booktabs}
\usepackage{subcaption}
\usepackage{tabularx}
\usepackage{multirow}
\usepackage{url}

\newcommand{\methodname}{{\textsc{X-Reset}}}

\newcommand{\rev}[1]{#1}

\newcommand{\PD}[1]{}

\begin{document}

\title[\methodname: Scaling Object-Centric RL via Cross-Embodiment Resets]{\methodname: Scaling Object-Centric Reinforcement Learning via Cross-Embodiment Resets}

\author{Prithwish Dan}
\affiliation{\institution{Applied Intuition}\city{Mountain View}\state{CA}\country{USA}}
\author{Chenyang Ma}
\affiliation{\institution{Applied Intuition}\city{Mountain View}\state{CA}\country{USA}}
\author{Wei Zhan}
\affiliation{\institution{Applied Intuition}\city{Mountain View}\state{CA}\country{USA}}

\compactauthors{%
  {\raggedright
   {\rmfamily\bfseries
    Prithwish Dan\textsuperscript{1}, \;
    Chenyang Ma\textsuperscript{1,2,*}, \;
    Wei Zhan\textsuperscript{1,3,\dag}\par}
   \smallskip
   {\small\rmfamily\color{AppliedGray}%
    \textsuperscript{1}Applied Intuition \quad
    \textsuperscript{2}University of North Carolina at Chapel Hill
    \quad
    \textsuperscript{3}University of California, Berkeley\par}
   \smallskip

   {\footnotesize\rmfamily\color{AppliedGray}%
 \textsuperscript{*}Work done during internship at Applied Intuition. \quad
 \textsuperscript{\dag}Corresponding author: wei.zhan@applied.co\par}

   \smallskip
   {\footnotesize\rmfamily\color{AppliedGray}%
    Project Website: \href{https://xreset-applied.github.io/}{https://xreset-applied.github.io/}\par}
   \par}
}

\begin{abstract}
Reinforcement learning (RL) in simulation can train dexterous manipulation policies without robot demonstrations, but training a single generalist policy with task-agnostic rewards faces a severe exploration problem: approaching, grasping, and reorienting diverse objects with many degrees of freedom is difficult to discover from scratch. Prior works make exploration tractable with high-quality robot demonstrations, per-task reward shaping, or by restricting policies to narrow modes of behavior. We propose \methodname, a framework that instead resolves exploration with human hand-object demonstrations. Rather than imitating or tracking retargeted human motion, \methodname\ kinematically retargets hand-object states to noisy robot states, filters out states that are unstable in simulation, and samples the remainder as resets during RL training with general-purpose object-centric rewards. The resulting policy depends only on object state and goal, with demonstrations entering training through the reset distribution. We show that \methodname\ trains generalist policies on 20 objects across three embodiments---a 22-DoF hand on two different arms and a parallel-jaw gripper---and resolves the exploration challenges of RL from scratch. \methodname\ scales with the number of training objects, generalizes to unseen objects, can learn from imperfect hand-pose estimates, and transfers behaviors zero-shot from sim-to-real.
\end{abstract}

\maketitle

\clearpage
\tableofcontents
\clearpage

\section{Introduction}

The predominant recipe for training robot manipulation policies is imitation learning (IL), in which robot data is used as supervision to learn a mapping from observations to actions~\cite{chi2023diffusionpolicy, pi05_2025}. Scaling such approaches to general-purpose models faces the fundamental bottleneck of requiring large-scale datasets of high-quality expert demonstrations~\cite{lin2024data}. Despite a multitude of collection systems~\cite{chi2024umi, handa2020dexpilot, chen2025dexforce, wang2024dexcap, arunachalam2023holodex, zhao2023learning, iyer2024openteach, yang2024ace}, teleoperation remains most expensive precisely where dexterity matters most: multi-fingered hands with many degrees of freedom (DoF). Reinforcement learning (RL) in simulation offers an alternative that requires no robot demonstrations, and has produced dexterous skills that transfer to the real world~\cite{openai2019rubiks, handa2023dextreme, chen2023visualdex}.

However, RL faces a large exploration burden when learning to approach, grasp, and reorient objects with many DoF from scratch, and prior works make the problem tractable with per-task scaffolding or restricted behavior. Demonstration-led curricula~\cite{demostart, tao2024rfcl} and favorable reset distributions~\cite{omnireset} reduce this burden but require high-quality robot demonstrations or hand-designed per-task states. Recent works that train generalist object-centric RL controllers instead mitigate exploration by learning only narrow modes of behavior, such as operating only on tool handles~\cite{simtoolreal} or strictly learning top-down power grasps via reward shaping~\cite{kuang2026dex4d}. We instead aim to train a single generalist policy to learn these skills on a diverse set of objects with a task-agnostic reward.

Human videos, which are abundant and encompass a diverse range of day-to-day hand-object interactions~\cite{damen2022epic, hoque2025egodex, chao2021dexycb}, are a natural source of guidance for this exploration problem. Many works use human data directly for imitation by kinematically retargeting hand motion to robot actions~\cite{qin2022dexmv, lepert2025phantom, masquerade}, but retargeting is only a rough estimate of robot motion and often fails to obey the true contact dynamics of robot-object interaction. Works that use RL to bridge the embodiment gap either transfer human object motion to train single-task policies~\cite{dan2025xsim, lum2025h2s2r, chen2024object} or use retargeted motion as a reference for RL~\cite{guzey2024hudor, zhu2026learningdexterousmanipulationusing, dexmachina, li2025maniptrans}, often relying on high-quality MoCap data and restricting learned behaviors to specifically demonstrated strategies. \textbf{\textit{Our key insight is that kinematically retargeted human hand-object states can supplement RL with task-agnostic rewards by offering a natural reset distribution of states from which robots can easily explore and access success signals during training.}} Resetting to reference states is standard practice in humanoid motion tracking~\cite{peng2018deepmimic, luo2023phc}, where the policy is conditioned on and rewarded for following the reference. For general-purpose manipulation, direct reference-tracking is under-specified under important randomizations such as novel goal poses and external forces that may knock objects into unseen states~\cite{lum2026play2perfect}; our policy instead depends only on object state and goal, and demonstrations enter training only through resets.

\begin{figure}[t]
\centering
\includegraphics[width=\textwidth]{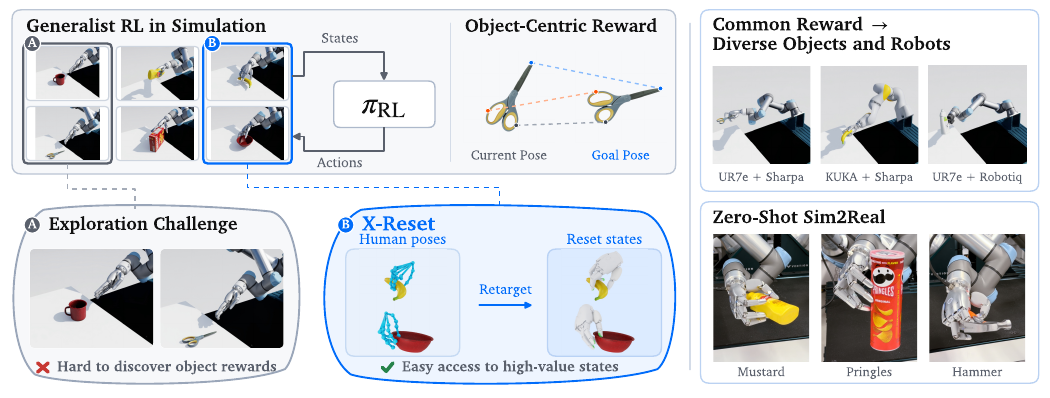}
\caption{\textbf{Overview of \methodname.} We introduce \methodname, a simple recipe to pair generalist object-centric RL in simulation with kinematically retargeted human hand-object states as resets. Learning to approach, grasp, and reorient objects from scratch results in a challenging exploration problem, while \methodname\ directly exposes policies to high-value states from which exploration is easy. \rev{With a common reward formulation, we train generalist policies for diverse robot embodiments and demonstrate sim-to-real transfer.}}
\label{fig:teaser}
\end{figure}

We propose \methodname, a framework to train dexterous manipulation policies by defining reset distributions over human hand-object states during RL training (Fig.~\ref{fig:teaser}). \methodname\ kinematically retargets hand-object trajectories to the target embodiment, filters out states that are unstable in simulation, and randomly resets to the retained states while training object-centric RL to approach, grasp, and reorient objects. These resets allow quicker access to success signals that are otherwise difficult to discover from scratch. Our contributions are summarized as follows:

\begin{enumerate}[leftmargin=*]
\itemsep0em
\item We propose \methodname, a framework that makes task-agnostic object-centric RL tractable for dexterous manipulation by framing human hand-object states as a reset distribution to remedy the exploration problem.
\item We show that \methodname\ trains generalist RL policies on 20 DexYCB~\cite{chao2021dexycb} objects with task-agnostic rewards. Across three embodiments---a five-fingered hand on two different arms and a parallel-jaw gripper---average reorientation success increases from $11.0\%$ to $66.6\%$ with object-only rewards, and from $42.1\%$ to $59.6\%$ with hand-object rewards.
\item We show that \methodname\ scales with the number of training objects, generalizes zero-shot to unseen objects, and serves as a base policy for fine-tuning on novel objects without additional human demonstrations. \methodname\ is also robust to offline hand-pose estimation errors at training time, and learned behaviors transfer zero-shot from simulation to the real world for a 22-DoF hand mounted on a 6-DoF arm on both seen and unseen objects.
\end{enumerate}

\section{Related Work}

\label{sec:related}



\textbf{Dexterous Manipulation with RL.} RL in simulation has produced dexterous skills that transfer to the real world~\cite{openai2019rubiks, handa2023dextreme, chen2023visualdex}, but exploration is a major bottleneck when training high-DoF hands from scratch. Demonstration-based initialization and curricula reduce this burden~\cite{florensa2017reverse, demostart, tao2024rfcl} but need high-quality robot data; OmniReset~\cite{omnireset} instead trains single-task policies from manually specified reset distributions for parallel-jaw grippers. Generalist controllers also exploit structured priors: \cite{simtoolreal, lum2026play2perfect} target handle-like tools using a shared grasp-region representation and reports that PPO alone is insufficient for exploration, while \cite{kuang2026dex4d} combines object-keypoint rewards with finger-curl shaping and learns only top-down power grasps. \methodname\ retains a generic object-centric reward and instead addresses exploration through the reset distribution, sourced from diverse human data rather than expensive robot data, and can learn behaviors on diverse objects ranging from scissors to bowls to powerdrills.

\textbf{Reference State Initialization.} Reference state initialization is standard in motion tracking for humanoids~\cite{peng2018deepmimic, luo2023phc} where the training objective explicitly rewards reference-following. Works in manipulation~\cite{li2025maniptrans, xu2025dexplore} use kinematic imitation rewards to shape exploration and reference-following, relying on MoCap data for accurate hand-object pose and contact estimation~\cite{dexmachina, li2025maniptrans, xu2025dexplore}. Reference-tracking for contact-rich manipulation faces diversity challenges, since there is no natural hand-object reference to imitate under important randomizations~\cite{lum2026play2perfect} such as novel goal poses or external forces which may knock objects over. Instead, our approach does not require tracking the reference at every step; it injects noisy retargeted resets to supplement object-centric RL training of policies which simply consume current object state and goal.

\textbf{Learning from Human Hand-Object Data.} Human videos offer a scalable source of object-interaction priors~\cite{grauman2022ego4d,  hoque2025egodex, chao2021dexycb}, but lack the action labels needed for standard IL. To bridge this, prior work maps human motion to robot actions through kinematic retargeting~\cite{qin2022dexmv, shaw2022videodex, li2024okami,qiu2025humanoid}, narrows the visual gap by overlaying rendered robot arms on human frames~\cite{lepert2025phantom, masquerade, shadow} or unifying both into a shared keypoint representation~\cite{ren2025motion_tracks, haldar2025point_policy, pace2025xdiffusion}, or decouples video-derived planning from low-level control~\cite{li2024okami, wang2023mimicplay, shi2025zeromimic}. These methods assume retargeted hand motions can be directly imitated on the robot, but this often fails due to embodiment mismatch and the true contact dynamics of robot-object interaction~\cite{dan2025xsim}, and they typically require co-training with substantial robot data to recover. Human-guided RL instead uses residual actions or motion/contact references~\cite{guzey2024hudor, zhu2026learningdexterousmanipulationusing,dexmachina, li2025maniptrans, xu2025dexplore}, or guides learning through object trajectories and hand-derived initialization~\cite{dan2025xsim, lum2025h2s2r, chen2024object}, typically producing single-task policies. \methodname\ uses human data neither as an imitation target nor as a per-task reference, but as a reset distribution for a single generalist policy.

\section{\rev{Problem Formulation}}
\label{sec:problem}

We focus on the problem of learning the core dexterous manipulation skills---approaching, grasping, and reorienting objects to goal poses---via object-centric RL in simulation, with reset states from diverse human demonstrations. For a given object $\eta$, a human hand-object trajectory $\xi^\eta = \{(h_t, s_t)\}_{t=1}^T$ consists of $T$ densely tracked timesteps of MANO~\cite{mano} hand poses where $h_t \in \mathbb{R}^{K\times3}$ represents the 3D position of each of the $K=21$ hand keypoints and $s_t \in \mathrm{SE(3)}$ represents the position and rotation of the object at timestep $t$. We define the goal pose $g = s_T$ as the last object pose from the human demonstration. In practice, we represent object poses by sampling $Z = 64$ keypoints on the object surface to enable a richer geometric representation: $s_t^{\rm kp}, g^{\rm kp} \in \mathbb{R}^{Z\times3}$. We denote a dataset of $N$ human hand-object trajectories $\mathcal{D}^\eta = \{\xi_i^{\eta}\}_{i=1}^N$, and construct a dataset of many objects $\mathcal{D} = \bigcup_{\eta} \mathcal{D}^\eta$. We learn a policy to produce actions \rev{$\mathbf{a}_t = \pi_{\phi}(q_t, s_t^{\rm kp}, g^{\rm kp}, m_t)$} given robot proprioception $q_t$, object keypoints $s_t^{\rm kp}$, goal keypoints $g^{\rm kp}$\rev{, and LSTM state $m_t$ encoding the interaction history; $\phi$ denotes the learned policy parameters}.
	
\section{Approach}
\label{sec:approach}
\begin{figure}[t]
  \centering
  \includegraphics[width=\textwidth]{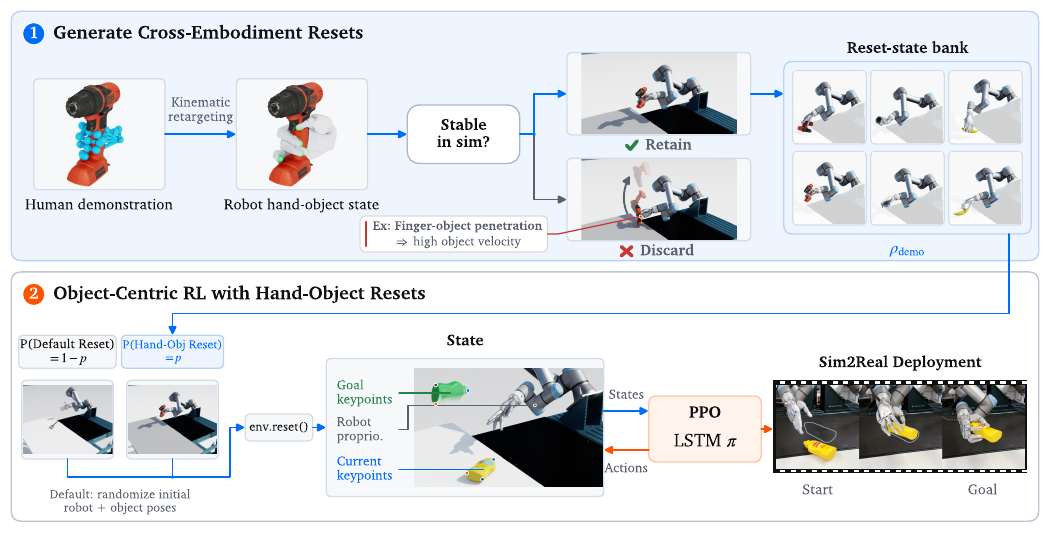}
  \caption{\textbf{\methodname\ Pipeline.} (Top) First, we kinematically retarget human hand-object states to noisy robot states. Then, we spawn states into simulation and retain a bank of stable states. (Bottom) During RL training, we randomly sample diverse resets leveraging the retargeted states, and train an RL policy which consumes object keypoints at inference time and can be deployed in the real world.}
  \label{fig:approach}
\end{figure}



We propose \methodname, which consists of two primary stages: (1) generating robot reset states from human hand-object poses; (2) training RL policies in simulation with object-centric rewards by incorporating random resets to demonstrated states to remedy the exploration problem.

\subsection{Generating Reset States from Human Hand-Object Demonstrations}\label{sec:generate_resets}

We first generate reset states for the robot from human demonstrations (Fig.~\ref{fig:approach}, Top). Such states must consist of (a) joint configurations for the robot, and (b) the 6D object pose \rev{and goal}, so that they can be spawned into simulation~\cite{isaac}. 

\textbf{Hand-Object Data Processing.}\label{approach:kr} We employ an optimization-based kinematic retargeting pipeline (using PyRoki~\cite{pyroki}) to estimate robot joint angles corresponding to each frame of the human demonstration. We first convert human hand pose $h_t \in \mathbb{R}^{K\times3}$ to robot hand joint angles $\theta_t^{\mathrm{hand}}$ using approximate fingertip and wrist keypoint targets following prior work~\cite{dexmachina}.  
Then, we perform collision-aware inverse kinematics (using cuRobo~\cite{sundaralingam2023curobo}) to solve for the remaining arm joints $\theta_t^{\mathrm{arm}}$ by tracking the wrist over time while keeping the hand poses fixed to finally obtain $\theta_t \in \mathbb{R}^{J}$ where $J$ is the number of joints in the target embodiment. This process allows us to generate a dataset of $L$ hand-object pairs for the robot $\mathcal{D}_{\rm robot} = \{(\theta_l, s_l)\}_{l=1}^L$.

\textbf{Filtering Unstable Resets.} In practice, kinematic retargeting from human to robot hand can result in unrealistic states depending on the embodiment gap. To avoid sampling unstable simulation resets, we perform offline filtering to mark and discard hand-object states which are undesirable. These may be a byproduct of (i) failure to find robot joint configurations within joint limits or (ii) instability when spawned in simulation (e.g., due to hand-object penetration).
\label{approach:filter}


\subsection{Supplementing Object-Centric Reinforcement Learning with Simulation Resets}


Next, we perform RL training with generic object-centric reward functions to train robots to manipulate objects to goal poses (Fig.~\ref{fig:approach}, Bottom). Key to our approach is the use of simulation resets: over the course of training, we occasionally reset to hand-object states generated in Section~\ref{sec:generate_resets}. By exposing the policy to diverse resets that lie in high-value regions, success signals from later stages of manipulation propagate and help avoid the collapse to local optima often faced by naive exploration from scratch.

\textbf{Object-Centric Reward Function.}\label{approach:reward}
We adopt a standard object-centric reward function for training RL policies which is common across all objects and robot policies:
\begin{equation}
\begin{split}
r_t = {}& r_{\rm prox}(q_t,s_t^{\rm kp})
+\mathds{1}_{\rm prox}r_{\rm reorient}(s_t^{\rm kp},g^{\rm kp}) \\
&+\mathds{1}_{\rm prox}r_{\rm success}(s_t^{\rm kp},g^{\rm kp})
+r_{\rm smooth}(\mathbf{a}_t)
+r_{\rm safety}
\end{split}
\end{equation}
The proximity reward $r_{\rm prox}$ encourages the robot hand to
approach the object, while the
reorientation reward $r_{\rm reorient}$ drives the object toward its goal pose.
$r_{\rm success}$ provides an additional bonus for achieving the goal pose
within an $\epsilon$ threshold, and $r_{\rm smooth}$ encourages smooth and
small actions. Finally, $r_{\rm safety}$ penalizes terminal failures such as
the object leaving the workspace or sustained contact between the robot and the table.

In practice, there are various metrics to measure the distance between the
current object pose $s_t$ and goal pose $g$ in order to compute
$r_{\rm reorient}$. Based on prior work~\cite{kuang2026dex4d, simtoolreal}, we \rev{compute} the mean distance
between corresponding keypoints:
\begin{equation}
d_{\rm kp}(s_t^{\rm kp},g^{\rm kp})
=
\frac{1}{Z}\sum_{z=1}^{Z}
\left\|s_{t,z}^{\rm kp}-g_z^{\rm kp}\right\|_2,
\qquad
r_{\rm reorient}
\propto
-d_{\rm kp}
\end{equation}
This distance jointly captures differences in object position and orientation.

\textbf{Resetting to Hand-Object States.}
\label{approach:reset} \rev{At initialization, each environment is assigned an object and a demonstration for that object.} We define the default reset distribution of the robot and object as $\rho_{\rm default}$ (including randomizations in start and goal poses), and $\rho_{\rm demo}$ as a uniform distribution over \rev{the retained frames of its assigned demonstration, using the corresponding retargeted hand-object states from $\mathcal{D}_{\rm robot}$}. At train time, we sample resets from $\rho_{\rm default}$ with probability $1-p$ and sample from $\rho_{\rm demo}$ with probability $p$. This reset scheme helps expose the policy $\pi_\phi$ to a diverse set of states in simulation which lie along noisy robot trajectories and can enable discovery of success signals otherwise inaccessible via naive exploration from $\rho_{\rm default}$.

\textbf{Policy Details.} We train our policy using PPO~\cite{schulman2017proximal}. Our policy consumes the current and goal object keypoints and robot proprioception. The LSTM architecture consumes the interaction history and enables adaptive behaviors. We also leverage an asymmetric critic~\cite{pinto2018asymmetric} that has access to additional privileged information to further stabilize training. We apply standard domain randomization and object/robot perturbations inspired by prior work~\cite{simtoolreal, lum2026play2perfect}. The policy can be deployed directly in the real world using vision foundation models to track object pose (see Section~\ref{experiments:realworld}).



\section{Experiments}
\label{sec:experiments}
\begin{figure}[t]
  \centering
  \includegraphics[width=\textwidth]{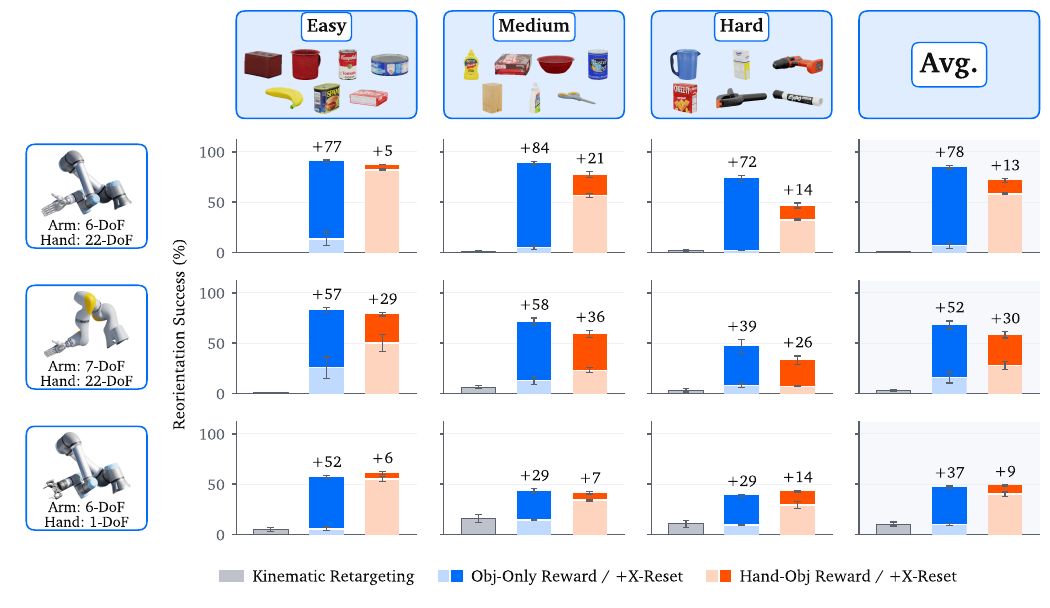}
  \caption{\textbf{\methodname\ Supplements RL Training.} We report reorientation success at $\epsilon = $2\,cm on 20 DexYCB objects (7 Easy, 7 Medium, 6 Hard). \rev{Evaluation uses 20 demonstrated start/goal pairs per object. Pale bars show RL baselines trained from scratch; saturated caps add the \methodname\ gain ($p=0.9$), with gain labels in percentage points.} Avg. weights all objects equally. \rev{\methodname\ improves both reward formulations across embodiments and object groups, with the best recipe being low-bias object-only rewards seeded with \methodname\ for exploration on average.} Error bars show 3-seed RL training SE.}
  \label{fig:barplots}
\end{figure}

\textbf{Experimental Setup.} \rev{We evaluate three embodiments: a 22-DoF Sharpa right hand on a 6-DoF UR7e arm, the same hand on a 7-DoF KUKA iiwa 14 arm, and a 1-DoF Robotiq 2F-85 parallel-jaw gripper on a UR7e arm. We train a separate generalist policy for each embodiment.} We use hand-object trajectories from DexYCB~\cite{chao2021dexycb} and IsaacSim~\cite{isaac} for physics simulation. Unless otherwise specified, each policy is jointly trained on 20 objects with 20 right-handed demonstrations per object, and \methodname\ uses $p=0.9$ as the probability of sampling hand-object resets from $\rho_{\rm demo}$. We train every simulation condition with three seeds.

\textbf{Evaluation Metrics.} In simulation, our primary task metric is \textit{Reorientation Success (\%)}, which measures whether the object reaches its goal pose within a keypoint error of $\epsilon=2\,\mathrm{cm}$ under the distance metric $d_{\rm kp}$ in Section~\ref{approach:reward}. We additionally measure \textit{Avg. Episode Return} from $\rho_{\rm default}$ to compare training dynamics across policies on a common state distribution. Simulation results report means over three seeds, with standard errors shown in the figures. We aim to answer the following questions:

\begin{enumerate}[leftmargin=*]
\itemsep0em
    \item \textbf{Cross-Embodiment Resets:} Does \methodname\ resolve the exploration problem faced by object-centric RL from scratch, and how does it compare against baselines on functionally retargeting human demonstrations?
    \item \textbf{\rev{Scaling and Generalization:}} What are \methodname's scaling and generalization properties when manipulating unseen objects without human demonstrations?
    \item \textbf{Reset Distribution:} How does the probability of sampling hand-object states affect training behavior?
    \item \textbf{Real-World Capabilities:} Can \methodname\ learn from noisy offline hand tracking and transfer behaviors from sim-to-real under object-tracking noise?
\end{enumerate}

\subsection{Cross-Embodiment Resets}
\label{exp:cross_embodiment}
We evaluate whether \methodname's hand-object resets make object-centric RL tractable across embodiments. \rev{This comparison uses the demonstrated object start/goal pairs from DexYCB (20 per object). We compare against direct kinematic playback and two dominant reward-family baselines.}
\begin{itemize}[leftmargin=*]
\itemsep0em
    \item \textbf{Kinematic Retargeting:} Used as the foundational backbone for prior works that perform Imitation Learning (IL) from human hand actions~\cite{masquerade, haldar2025point_policy, ren2025motion_tracks}, this baseline defines a kinematic mapping between 3D MANO hand positions~\cite{mano} and robot joints. Robot arm and hand joint angles are then solved for via Inverse Kinematics (IK) (Section~\ref{approach:kr}), and rolled out \textit{open-loop}.
    \item \textbf{Obj-Only Reward:} The generic object-keypoint reward of Section~\ref{approach:reward}, following the task-agnostic formulation of generalist controllers~\cite{simtoolreal, kuang2026dex4d}. $r_{\text{reorient}}$ drives the object toward its goal and $r_{\text{prox}}$ encourages the hand to approach the object surface.
    \item \textbf{Hand-Obj Reward:} Adds hand guidance inspired by reference-tracking works~\cite{dexmachina, li2025maniptrans, xu2025dexplore}. It shapes $r_{\text{prox}}$ to encourage robot fingertips to lie near the human fingertips at the demonstrated goal pose, and remains well-defined under perturbed object states by anchoring relative to the object itself.
\end{itemize}

 We group the 20 objects into 7 \textit{Easy}, 7 \textit{Medium}, and 6 \textit{Hard} objects based on learnability for baselines, trending from simple to complex geometries (Fig.~\ref{fig:barplots}). \textbf{Kinematic Retargeting} struggles to establish stable grasps due to the human-robot embodiment gap, with an average reorientation success of only $4.9\%$ across the three embodiments. \textbf{Obj-Only Reward} struggles with naive exploration, reaching $15.0\%$ average success on \textit{Easy} objects and $11.0\%$ overall. \rev{\textbf{Hand-Obj Reward} benefits from hand-guided shaping and reaches $42.1\%$ average success overall.}

Applying \methodname\ to both reward structures improves performance across all three embodiments (Fig.~\ref{fig:barplots}). \rev{Averaged over objects and embodiments, reorientation success increases from $11.0\%$ to $66.6\%$ for Obj-Only Reward and from $42.1\%$ to $59.6\%$ for Hand-Obj Reward. \textbf{Obj-Only Reward+\methodname} achieves a higher success rate despite the lower starting performance of its from-scratch baseline.} These results support the general recipe: \methodname\ to seed exploration, and  Obj-Only Rewards with minimal bias and high freedom to explore robust embodiment-specific strategies instead of explicitly rewarding closeness to the human demos.

\subsection{\rev{Scaling and Generalization}}
\label{exp:scaling}
\begin{figure}[t]
  \centering
  \includegraphics[width=\textwidth]{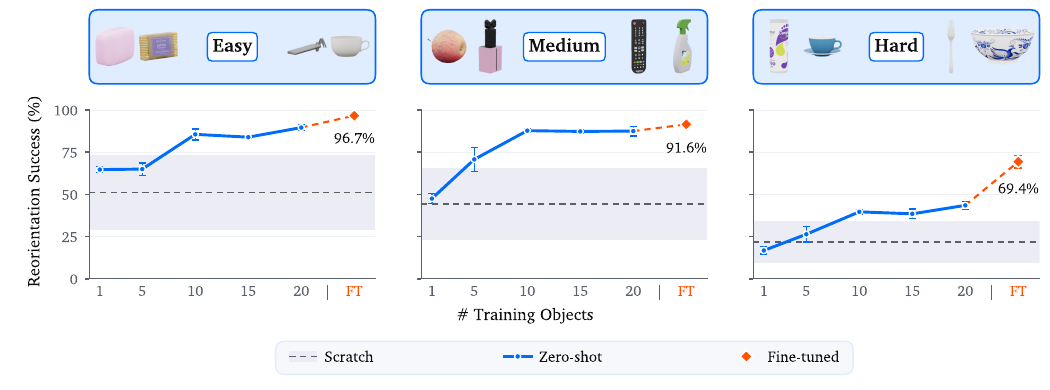}
  \caption{\textbf{\rev{Scaling and Generalization.}} We report reorientation success at $\epsilon=$ 2\,cm on 12 EBench objects. Scratch baseline trains on the EBench roster for its full 10k epochs. Blue connected markers show zero-shot transfer after 10k epochs of \methodname\ pretraining on 1--20 DexYCB objects, outperforming training from scratch on EBench with as little as 10 DexYCB training objects. Orange diamonds (FT) show 5k epochs of 20-object \methodname\ pretraining followed by 5k epochs of EBench fine-tuning, showing \methodname\ is a strong pre-training foundation. Error bars show 3-seed RL training SE.}
  \label{fig:scaling}
\end{figure}

We evaluate \textbf{Obj-Only Reward+\methodname}'s ability to transfer to unseen objects as we vary the number of training objects. We train UR7e+Sharpa policies on a range of 1-20 DexYCB objects and evaluate zero-shot reorientation on a fixed set of 12 objects from EBench~\cite{ebench_assets}, with four objects per difficulty category (Fig.~\ref{fig:scaling}). \rev{We evaluate each object over 400 trials with varying start/goal conditions.} Increasing training diversity improves transfer overall: on \textit{Easy} objects, which share more familiar geometries with the training data, success starts high and rises from $64.8\%$ to $89.6\%$. On \textit{Hard} objects, including a very large bowl and a thin fork, we improve from $16.8\%$ to $43.5\%$.

\rev{We also study \methodname's adaptability under a matched 10k-epoch budget: 5k epochs of DexYCB pretraining followed by 5k EBench epochs (Fine-tuned), versus 10k EBench epochs from random initialization (Scratch). Both policies train on the same 12 EBench objects. Because the assets have no corresponding human demonstrations, both EBench training segments use default resets ($p=0$).} Fine-tuning reaches $85.9\%$ average reorientation success, compared to just $39.1\%$ from scratch. \methodname\ therefore also serves as an initialization for adapting to novel objects without new human demonstrations.

\subsection{Reset Distribution}
\label{exp:reset_distribution}

\begin{figure}[t]
  \centering
  \includegraphics[width=0.72\textwidth]{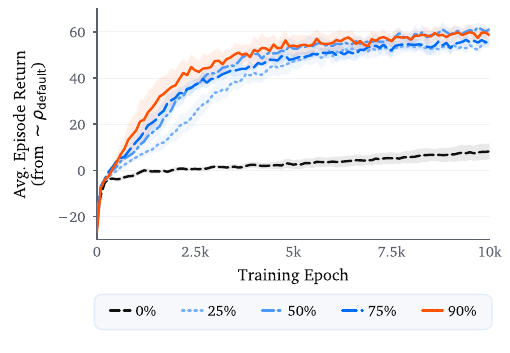}
  \caption{\textbf{Impact of Demo-Reset Probability.} We measure \textit{Avg. Episode Return} from $\rho_{\rm default}$ at train-time for UR7e+Sharpa with Obj-Only Reward. Higher demo-reset probability accelerates early learning, with all \methodname\ policies converging in a similar band. Curves show 3-seed means and shaded bands show $\pm1$ SE.}
  \label{fig:resetdist}
\end{figure}

We ablate \textbf{Obj-Only Reward+\methodname}'s training behavior on UR7e+Sharpa as we vary $p$, the probability of sampling resets from $\rho_{\rm demo}$ (Fig.~\ref{fig:resetdist}). We analyze \textit{Avg. Episode Return} from $\rho_{\rm default}$ to fairly compare all policies. Early in training, higher sampling rates of hand-object states accelerate learning. As training converges, \methodname\ policies reach returns between $55.45$ and $61.09$, all substantially outperforming the $0\%$ condition at $8.28$. The benefit therefore holds over a broad range of sampling probabilities, rather than depending on $p=0.9$ specifically.

\subsection{Real-World Capabilities}
\label{experiments:realworld}
\begin{figure}[t]
  \centering
  \includegraphics[width=0.62\textwidth]{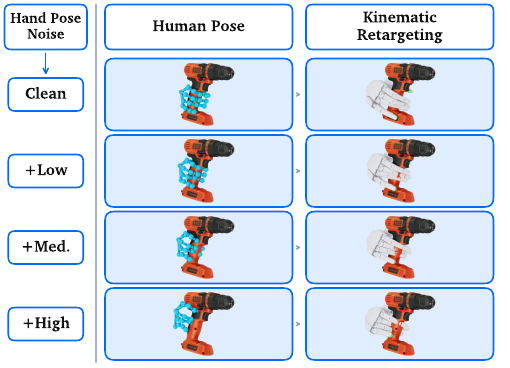}
  \par
  \begin{minipage}{\linewidth}
  \centering
  \footnotesize
  \setlength{\tabcolsep}{5pt}
  \renewcommand{\arraystretch}{1.15}
  \begin{tabular}{@{}clc@{}}
    \toprule
    Demo resets & Hand-pose noise & Avg. episode return \\
    \midrule
    $0\%$ & -- & $8.28 \pm 3.40$ \\
    \midrule
    $90\%$ & Clean & $58.53 \pm 0.93$ \\
    $90\%$ & Low & $56.07 \pm 4.57$ \\
    $90\%$ & Medium & $51.46 \pm 1.12$ \\
    $90\%$ & High & $53.29 \pm 2.58$ \\
    \bottomrule
  \end{tabular}
\end{minipage}

  \caption{\textbf{Learning from Noisy Hand Poses.} (Top) We visualize retargeted Sharpa hand-object states under varying levels of noise. \rev{(Bottom) Avg. episode return from $\rho_{\rm default}$ of \methodname\ trained with noisy hand poses.}}
  \label{fig:pose_noise}
\end{figure}

\textbf{Learning from Noisy Hand Poses.} Since DexYCB estimates hand and object pose from privileged multi-view cameras, we inject varying levels of noise into the hand poses to test whether policies still learn from imperfect offline data. \rev{We add random wrist noise that translates the whole hand, plus per-keypoint finger noise}. We perturb the same clean hand poses at each noise level, then apply kinematic retargeting and the instability filtering in Section~\ref{approach:filter}. Fig.~\ref{fig:pose_noise} shows the noisy human poses, corresponding robot configurations, and \textit{Avg. Episode Return} for UR7e+Sharpa with Obj-Only Reward. Compared to a return of $58.53$ with clean poses, the noisy conditions reach $51.46$--$56.07$, again well above the $8.28$ obtained from exploration without demonstration resets. \rev{The High-noise condition also has a $2.45\times$ higher reset rejection rate than Clean (6.82\% to 16.71\%), which may explain some of the degradation in performance.} Still, a generic object-centric reward is useful here: noisy hand estimates enter training only through resets, whereas in reference-tracking methods~\cite{dexmachina, li2025maniptrans, xu2025dexplore} the same noise would corrupt the imitation target itself.

\begin{figure}[t]
  \centering
  \includegraphics[width=0.72\textwidth]{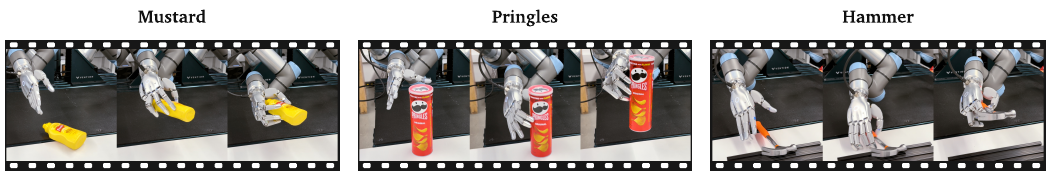}
  \par\smallskip
    \footnotesize
  \setlength{\tabcolsep}{4pt}
  \renewcommand{\arraystretch}{1.15}
  \begin{tabular}{@{}lccc@{}}
    \toprule
     & In-Distribution & \multicolumn{2}{c}{Out-of-Distribution} \\
    \cmidrule(lr){2-2}\cmidrule(l){3-4}
     & Mustard & Pringles & Hammer \\
    \midrule
    \rev{Grasp success} & \rev{8/10} & \rev{10/10} & \rev{8/10} \\
    \rev{Reorientation success} & \rev{8/10} & \rev{10/10} & \rev{6/10} \\
    \bottomrule
  \end{tabular}

  \caption{\textbf{\rev{Sim-to-Real Evaluation.}} (Top) UR7e+Sharpa filmstrips show each object manipulated to goal poses in the air. \rev{(Bottom) Grasp and reorientation successes out of 10 physical trials per object for Obj-Only Reward+\methodname.}}
  \label{fig:real_transfer}
\end{figure}

\textbf{Sim-to-Real Transfer.} \rev{The actor uses robot proprioception and object-centric observations available through a vision pipeline, while the asymmetric critic receives additional simulation information. We reconstruct object meshes with SAM 3D~\cite{sam3d3dfyimages}, track poses with FoundationPose~\cite{foundationpose}, and sample surface keypoints. We deploy the UR7e+Sharpa policy on seen (Mustard $\in$ DexYCB) and unseen (Pringles, Hammer $\notin$ DexYCB) on a range of start and goal poses. The policy successfully transfers behaviors to the real hardware; the table in Fig.~\ref{fig:real_transfer} reports the per-object results.} Qualitatively, the policy also shows recovery behavior and robustness to perturbation, for example sliding fingers underneath the Mustard when it is slipping (Fig.~\ref{fig:real_transfer}, Top Left). \rev{The primary observed failure mode is object-pose estimation error, motivating future visual or visuo-tactile policy distillation.}




\section{Discussion and Limitations}
\label{sec:discussion}

We present \methodname, which trains generalist dexterous manipulation policies by using retargeted human hand-object states as a reset distribution for object-centric RL. Resets make task-agnostic rewards tractable where exploration from scratch fails, and the resulting policies scale with training objects, tolerate noisy hand tracking, and transfer zero-shot to hardware.

\textbf{Limitations.} Consuming object pose makes the policy inherit tracking noise; visual~\cite{dan2025xsim, omnireset} or visuo-tactile distillation~\cite{lee2026adept} is a natural remedy, and can be combined with high-level planners to execute long-horizon as a sequence of goal-reaching subtasks~\cite{simtoolreal, kuang2026dex4d}. We also have yet to scale to large egocentric datasets~\cite{hoque2025egodex, damen2022epic}, which would require real-to-sim pipelines that extract hand-object states from video for rigid, articulated, and deformable objects.

\bibliographystyle{ACM-Reference-Format}
\bibliography{main_v2}

\newpage
\appendix
\begingroup
\section{Implementation and Evaluation Details}
\setcounter{equation}{0}
\renewcommand{\theequation}{A\arabic{equation}}
\renewcommand{\theHequation}{appendix.\arabic{equation}}
\setcounter{table}{0}
\renewcommand{\thetable}{A\arabic{table}}
\renewcommand{\theHtable}{appendix.\arabic{table}}
\setlength{\tabcolsep}{3pt}

\subsection{Data and Representation}
\label{app:representation}
DexYCB~\cite{chao2021dexycb, mano} supplies synchronized MANO keypoints, object poses, and meshes. A common table-aligned transform maps each demonstration's hand and object into simulation; robot-base transforms map wrist targets into arm coordinates. We sample 64 ordered object-local surface points $u_z$ (collision surfaces for DexYCB; visual surfaces for EBench):
\begin{equation}
\begin{gathered}
 s_{t,z}^{\rm kp}=R_tu_z+p_t,\qquad g_z^{\rm kp}=R_gu_z+p_g,\\
 f_{t,z}=[s_{t,z}^{\rm kp},\ g_z^{\rm kp},\ g_z^{\rm kp}-s_{t,z}^{\rm kp}]\in\mathbb R^9.
\end{gathered}
\end{equation}
Point indices define current/goal correspondence. The actor receives these paired features; its LSTM supplies temporal context.

\subsection{Retargeting, Filtering, and Resets}
\label{app:retargeting}
\label{app:filter}
\label{app:reset}
\begin{itemize}[leftmargin=*,nosep]
\item \textbf{Retargeting:} PyRoki jointly optimizes Sharpa hand joints and root poses over each sequence. Robotiq uses an analytical hand-to-gripper mapping. cuRobo solves collision-aware arm IK with the hand fixed, including the robot and table/workstation geometry and a home-to-first-target approach segment.
\item \textbf{Contact objective:} squared residuals $e_{t,j}=5m_{t,j}[|c_{t,j}-x_{t,j}|-0.01]_+$, with componentwise deadband in meters and validity mask $m_{t,j}$. Other residual multipliers are listed below.
\item \textbf{Filtering:} we hold IK-valid configurations at their initial joint targets with gravity disabled, and reject states that terminate, produce non-finite measurements, or violate the thresholds below. Final gates test the last sample; peak gates test the entire window. Both Sharpa embodiments share settings.
\item \textbf{Sampling:} each environment retains an assigned object/demo. With probability $p$, sample uniformly among its retained frames and use that frame's robot/object state with the demonstration's terminal object goal. Otherwise use the home robot configuration and demonstrated object start. Human-derived training uses $p=0.9$ unless ablated; synthetic EBench training uses $p=0$.
\end{itemize}

\begin{center}
\footnotesize
\captionof{table}{Retargeting and reset-filter settings.}
\label{tab:appendix_filter}
\begin{tabular}{@{}p{0.55\columnwidth}p{0.39\columnwidth}@{}}
\toprule
Retargeting parameter & Value \\
\midrule
Local / global alignment multiplier & 10 / 1 \\
Joint / root-translation smoothness & 2 / 2 \\
Joint-limit / rest-pose multiplier & 100 / 0.2 \\
Arm IK seeds & 64 \\
Wrist position / rotation tolerance & 5\,mm / 0.05\,rad \\
\midrule
Reset-filter parameter & Sharpa / Robotiq \\
\midrule
Hold-target steps (30\,Hz) & 24 / 120 \\
Final object linear speed (m/s) & 0.5 / 0.5 \\
Final object angular speed (rad/s) & 15 / 15 \\
Final maximum joint speed (rad/s) & 10 / 10 \\
Peak object linear speed (m/s) & Off / 1.0 \\
Peak gripper-joint speed (rad/s) & Off / 2.0 \\
\bottomrule
\end{tabular}
\end{center}

\subsection{Rewards}
\label{app:reward}
We adopt several reward terms from~\cite{kuang2026dex4d}. Let $d=d_{\rm kp}(s_t^{\rm kp},g^{\rm kp})$, $P$ be the sum of fingertip distances to the nearest object surface keypoints, and $M$ the analogous palm distance (all in meters). The object-only reward terms are
\begin{equation}
\begin{aligned}
 r_{\rm prox}&=-2.5\min(P,3)-2.5\min(M,0.5),\\
 G&=\mathds{1}\{P\leq0.6\ \land\ M\leq0.2\},\\
 r_{\rm reorient}&=3(1.4-3d),\\
 r_{\rm success}&=\frac{15}{1+10d}\mathds{1}\{d\leq0.02\}.
\end{aligned}
\end{equation}
The geometric gate $G=\mathds{1}_{\rm prox}$ multiplies the two goal terms. For clipped actions $\bar{\mathbf a}_t\in[-1,1]^{n_a}$ and $\Delta\bar{\mathbf a}_t=\bar{\mathbf a}_t-\bar{\mathbf a}_{t-1}$,
\begin{equation}
\begin{aligned}
 r_{\rm smooth}={}&-0.005\|\bar{\mathbf a}_t\|_2^2
 -0.05\|\Delta\bar{\mathbf a}^{\rm arm}_t\|_2^2\\
 &-0.005\|\Delta\bar{\mathbf a}^{\rm hand}_t\|_2^2.
\end{aligned}
\end{equation}
Safety penalties are $-500$ each for an out-of-bounds object and terminal robot--environment contact (any configured arm/hand sensor above 5\,N for 0.5\,s). No dense contact penalty is used. The environment multiplies weighted rewards by $\Delta t=1/30$\,s; PPO additionally scales rewards by 0.01, unlike the reported environment return.

\textbf{Hand-object variant.} Object-local fingertip targets $c_f$ come from the last valid IK frame and follow the current object pose, $\tilde c_{t,f}=R_tc_f+p_t$. Replace $r_{\rm prox}$ by $-2.5\min(\sum_f\|x_{t,f}-\tilde c_{t,f}\|_2,3)$ and $G$ by a mean fingertip-to-target distance threshold of 0.05\,m. Other terms are unchanged; targets specify a grasp region, not a time-indexed trajectory.

\subsection{Policy and Training}
\label{app:policy}
The actor observes joint positions, controller targets, fingertip-to-object geometry, and paired point features. The critic additionally observes clean geometry, joint/object velocities, goal error, closest-fingertip information, contacts, reward state, and perturbations. UR7e+Sharpa has 648 pre-encoder observations and 28 actions; dimensions vary by embodiment.

\begin{center}
\footnotesize
\captionof{table}{Architecture and PPO. Batch quantities are per GPU rank.}
\label{tab:appendix_ppo}
\begin{tabular}{@{}p{0.46\columnwidth}p{0.48\columnwidth}@{}}
\toprule
Setting & Value \\
\midrule
PointNet MLP / pooling & $9\!\to\!64\!\to\!128\!\to\!128$ / max \\
Actor LSTM / ELU MLP & 1 layer, 512 units, layer norm / $[512,256,128]$ \\
Critic LSTM / ELU MLP & 512 units / $[512,512,256]$ \\
Action head & Gaussian; state-independent log std. \\
Implementation & RL Games PPO \\
Discount / GAE / PPO clip & 0.99 / 0.95 / 0.2 \\
Learning rate / KL target & $10^{-3}$ adaptive / 0.01 \\
Entropy / value / bounds loss & 0 / 4 / $10^{-4}$ \\
Reward scale / gradient cap & 0.01 / 1.0 \\
Rollout / recurrent length & 32 / 32 \\
Mini-epochs / minibatch & 5 / 32,768 \\
Environments / transitions per update & 8,192 / 262,144 \\
Input / value / advantage normalization & Enabled \\
Observation history / episode limit & 1 / 12\,s \\
Physics / policy frequency & 120 / 30\,Hz \\
\bottomrule
\end{tabular}
\end{center}

\textbf{Control.} Actions increment persistent joint-position targets. UR7e arm increment/velocity-limit scales are 0.0125/0.2; Sharpa uses increment scale $1/24$, 90\% joint ranges, and effort scale 0.75. Other embodiments use their own controllers/output heads.

\textbf{Budgets.} Each seed (42/52/62) trains on a single GPU. Reported checkpoints are at 10k epochs. Fine-tuning uses 5k DexYCB + 5k EBench epochs; scratch uses 10k EBench epochs, matched within each seed.

\begin{center}
\footnotesize
\captionof{table}{Training randomization.}
\label{tab:appendix_dr}
\begin{tabular}{@{}p{0.46\columnwidth}p{0.48\columnwidth}@{}}
\toprule
Quantity & Range / setting \\
\midrule
Default object XY / yaw & $\pm0.10$\,m / $\pm25^\circ$ \\
Default arm / hand joint noise & $\sigma=0.075/0.10$\,rad \\
Default goal XY / yaw / Z & $\pm0.03$\,m / $\sim\!\pm30^\circ$ / $\pm0.05$\,m \\
Object/table friction & $[0.2,1.3]$; 250 buckets \\
Restitution & $[0,0.3]$ \\
Object / robot mass scale & $[0.7,1.3]$ / $[0.9,1.1]$ \\
Object center-of-mass offset & $\pm1$\,cm \\
Gain / extra arm damping scale & Log-uniform $[0.7,1.3]$ / $[0.8,1.2]$ \\
Actuator friction/armature/effort scale & $[0.7,1.3]$ \\
Sharpa elastomer / metal friction & $[0.4,1.2]$ / 0.1 \\
Arm / hand transport delay & 0--6 / 0--16 physics steps \\
Perception / proprioception delay & 0--6 / 0--1 policy steps \\
Pose translation / rotation noise & $\pm5$\,mm / $\pm3.75^\circ$ per correlated and uncorrelated component \\
\bottomrule
\end{tabular}
\end{center}

\textbf{Disturbances.} While the proximity gate is active, random-direction object forces satisfy $\|F\|\leq150m$\,N (mass $m$ in kg); arm pushes are bounded by 600\,N. Spin impulses use $|\Delta\omega|\leq30$\,rad/s, converted to torque using inertia and the physics timestep. Per-environment event probabilities are log-uniform in $[0.001,0.1]$.

\subsection{Evaluation}
\label{app:evaluation}
Success is $\mathds{1}\{\min_t d_{\rm kp}(s_t^{\rm kp},g^{\rm kp})\leq0.02\,\mathrm m\}$ over the rollout, independent of the reward gate and including initially near-goal cases. Episode returns use default resets. Simulation plots show mean $\pm$ SE over three training seeds, with $\mathrm{SE}=\mathrm{std}(\mu_1,\mu_2,\mu_3)/\sqrt3$ (sample std.). Object averages are computed within each seed before aggregation.

\begin{center}
\footnotesize
\captionof{table}{Evaluation protocols and cohorts.}
\begin{tabular}{@{}p{0.26\columnwidth}p{0.68\columnwidth}@{}}
\toprule
Study & Protocol \\
\midrule
DexYCB & 20 objects $\times$ 20 demonstrated start/goal pairs; 400 pairs per embodiment. Easy/Medium/Hard: 7/7/6; objects weighted equally. \\
EBench & 12 objects (4 per difficulty), used for both fine-tuning/scratch training and evaluation. Zero-shot policies exclude EBench training. \\
Goal families & Lift, $60^\circ$ yaw, $60^\circ$ x-tilt, $60^\circ$ y-tilt; 20 pairs each. \\
Trials & 80 pairs $\times$ 5 deterministic draws = 400/object; 4,800/policy/seed, 14,400 across seeds. \\
Start offsets & XY $\pm0.02$\,m; yaw $\pm25^\circ$; seed base 42. \\
Goal offsets & XY $\pm0.03$\,m; yaw $\sim\!\pm30^\circ$; Z $\pm0.05$\,m; seed base 71. \\
Physical & 10 attempts/object; grasp and reorientation counts use all attempts (Fig.~\ref{fig:real_transfer}). \\
\bottomrule
\end{tabular}
\end{center}
EBench Easy: two different soaps, bookmark, mug; Medium: apple, perfume, remote, detergent; Hard: can, mug, fork, bowl. Evaluation perturbation seeds are distinct from training seeds.

\subsection{Offline Hand-Pose Noise}
\label{app:noise}
Before retargeting/filtering, translate the whole hand by one uniformly sampled wrist offset per demonstration. Add palm-frame finger-keypoint noise with eight-frame correlation, $e_t=\rho e_{t-1}+\sqrt{1-\rho^2}\sigma\varepsilon_t$, $\rho=e^{-1/8}$, $\varepsilon_t\sim\mathcal N(0,I)$, and $e_0\sim\mathcal N(0,\sigma^2I)$. Clip vector norms, rotate to world coordinates, and restore bone lengths; caps apply before restoration. Object poses and untracked sentinel frames remain unchanged. Matched episode seeds are used across tiers.
\\
\begin{center}
\footnotesize
\captionof{table}{Hand-pose noise (mm). Wrist bounds are per coordinate.}
\label{tab:appendix_noise}
\begin{tabular}{@{}lccc@{}}
\toprule
Tier & Wrist bound & Finger $\sigma$ & Finger cap \\
\midrule
Clean & 0 & 0 & 0 \\
Low & 10 & 1.5 & 3 \\
Medium & 20 & 3 & 6 \\
High & 40 & 6 & 12 \\
\bottomrule
\end{tabular}
\end{center}
All tiers use the same retargeting/filter, object-only reward, $p=0.9$, and 10k-epoch budget; Clean reuses the clean-policy runs. The High tier rejects $2.45\times$ as many resets as Clean; corruption changes both reset quality and retained-bank composition.
\endgroup

\end{document}